\PassOptionsToPackage{unicode}{hyperref}
\PassOptionsToPackage{hyphens}{url}
\documentclass[
  11pt,
]{article}
\usepackage{amsmath,amssymb}
\usepackage{setspace}
\usepackage{iftex}
\ifPDFTeX
  \usepackage[T1]{fontenc}
  \usepackage[utf8]{inputenc}
  \usepackage{textcomp} 
\else 
  \usepackage{unicode-math} 
  \defaultfontfeatures{Scale=MatchLowercase}
  \defaultfontfeatures[\rmfamily]{Ligatures=TeX,Scale=1}
\fi
\usepackage{lmodern}
\ifPDFTeX\else
\fi
\IfFileExists{upquote.sty}{\usepackage{upquote}}{}
\IfFileExists{microtype.sty}{
  \usepackage[]{microtype}
  \UseMicrotypeSet[protrusion]{basicmath} 
}{}
\makeatletter
\@ifundefined{KOMAClassName}{
  \IfFileExists{parskip.sty}{%
    \usepackage{parskip}
  }{
    \setlength{\parindent}{0pt}
    \setlength{\parskip}{6pt plus 2pt minus 1pt}}
}{
  \KOMAoptions{parskip=half}}
\makeatother
\usepackage{xcolor}
\usepackage[margin=0.82in]{geometry}
\usepackage{color}
\usepackage{fancyvrb}

\DefineVerbatimEnvironment{Highlighting}{Verbatim}{commandchars=\\\{\}}
\newenvironment{Shaded}{}{}

\newcommand{\NormalTok}[1]{#1}

\usepackage{longtable,booktabs,array}
\usepackage{calc} 
\usepackage{etoolbox}
\makeatletter
\patchcmd\longtable{\par}{\if@noskipsec\mbox{}\fi\par}{}{}
\makeatother
\IfFileExists{footnotehyper.sty}{\usepackage{footnotehyper}}{\usepackage{footnote}}
\makesavenoteenv{longtable}
\usepackage{graphicx}
\makeatletter
\def\maxwidth{\ifdim\Gin@nat@width>\linewidth\linewidth\else\Gin@nat@width\fi}
\def\maxheight{\ifdim\Gin@nat@height>\textheight\textheight\else\Gin@nat@height\fi}
\makeatother
\setkeys{Gin}{width=\maxwidth,height=\maxheight,keepaspectratio}
\makeatletter
\def\fps@figure{htbp}
\makeatother
\NewDocumentCommand\citeproctext{}{}
\NewDocumentCommand\citeproc{mm}{%
  \begingroup\def\citeproctext{#2}\cite{#1}\endgroup}
\makeatletter
 \let\@cite@ofmt\@firstofone
 \def\@biblabel#1{}
 \def\@cite#1#2{{#1\if@tempswa , #2\fi}}
\makeatother
\newlength{\cslhangindent}
\newlength{\csllabelwidth}
\newenvironment{CSLReferences}[2] 
 {\begin{list}{}{%
  \setlength{\itemindent}{0pt}
  \setlength{\leftmargin}{0pt}
  \setlength{\parsep}{0pt}
  \ifodd #1
   \setlength{\leftmargin}{\cslhangindent}
   \setlength{\itemindent}{-1\cslhangindent}
  \fi
  \setlength{\itemsep}{#2\baselineskip}}}
 {\end{list}}
\usepackage{calc}

\ifLuaTeX
\usepackage[bidi=basic]{babel}
\else
\usepackage[bidi=default]{babel}
\fi
\babelprovide[main,import]{american}

\def\languageshorthands#1{}
\usepackage{booktabs,longtable,array,amsmath,amssymb,amsthm,graphicx,float}
\usepackage{microtype}
\usepackage{fancyhdr}
\usepackage{caption}
\usepackage{needspace}
\usepackage{fvextra}
\DefineVerbatimEnvironment{Highlighting}{Verbatim}{breaklines,commandchars=\\\{\}}
\makeatletter
\AtBeginDocument{\author{Guangzhe Zhang\\[3pt]\small Independent AI Researcher\\\small\texttt{twhite.zh@gmail.com}}}
\makeatother
\ifLuaTeX
  \usepackage{selnolig}  
\fi
\usepackage{bookmark}
\IfFileExists{xurl.sty}{\usepackage{xurl}}{} 
\hypersetup{
  pdftitle={Correct Now, Insufficient Later: Auditing Update Sufficiency in Context Compression},
  pdfauthor={Guangzhe Zhang},
  pdflang={en-US},
  hidelinks,
  pdfcreator={LaTeX via pandoc}}

\title{Correct Now, Insufficient Later: Auditing Update Sufficiency in
Context Compression}
\author{Guangzhe Zhang}
\date{17 September 2026}

\begin{document}
\maketitle

\setstretch{1.04}
\section*{Abstract}\label{abstract}
\addcontentsline{toc}{section}{Abstract}

A memory can answer a current query correctly while discarding
distinctions required by a later update. We investigate this failure
with a paired-history audit: two histories have the same current answer,
receive a shared future update, and require different subsequent
answers. A pilot evaluates 24 history pairs across six synthetic
mechanisms, 12 memory conditions, two repeats, and two model backends. A
deterministic frontier selector obtains strict reveal accuracy of 96/96
on DeepSeek and 82/96 on GLM; a structured writer obtains 62 successes
with one unresolved outcome and 56/96. The configured four-outcome joint
contrast has finite-sample identification intervals of {[}0.521,
0.542{]} and {[}0.292, 0.313{]}, not confidence intervals. A
record-level audit distinguishes retained-state adequacy, response
delivery, and answer-schema compliance without changing those original
scores. It finds 26 and 25 well-formed but semantically wrong structured
reveal memories, while all 14 GLM frontier reveal failures contain
correct values in the wrong wrapper. Tombstone removal produces 16/16
exact replay failures in the targeted mechanism. Identifier renaming
then exposes a separate flaw: original frontier late-reference adequacy
falls from 8/8 to 94/320 transformed instances. We provide and test a
label-equivariant repair, but it preserves only 2/8 original
late-reference answers: eliminating a naming shortcut does not solve
unknown future relevance. These results support a scoped evaluation
methodology and reproducible failure analysis, not general superiority
of the repaired algorithm. Paid pilot evidence, retrospective
diagnostics, and new offline tests are reported separately; no
independent held-out or natural-task validation is claimed.

\textbf{Keywords:} context compression; update sufficiency; agent
memory; temporal validity; metamorphic testing; reproducibility.

\section{1 Introduction}\label{introduction}

Context compression selects information for future use. A short memory
may accurately describe the present and still omit an older version, a
persistent revocation, or a dependency that becomes decisive later.
Consider two histories whose current value is X. Their preceding valid
values are A and B. After the same event revokes X, the correct answers
diverge. A memory that retained only X cannot recover both alternatives
without another source of information. Current-answer agreement is
therefore an inadequate stand-alone test of a memory's usefulness under
updates.

This observation is related to longstanding predictive-state reasoning,
which represents history through its implications for future
observations, rather than a new discovery that temporal information
matters (\citeproc{ref-littman2001predictive}{Littman, Sutton, and Singh
2001}). Contemporary memory evaluations also include temporal reasoning
and knowledge updates (\citeproc{ref-wu2024longmemeval}{Wu et al. 2024},
\citeproc{ref-wu2026longmemevalv2}{2026}). The narrower question here is
methodological: what can be learned by holding the current answer fixed,
applying the same future to different histories, and inspecting both the
stored evidence and the eventual response?

We study this question in CCA v4, a small symbolic event-log
environment. Its paired histories agree on the current query but differ
in latent evidence. A revealing future makes that evidence necessary; an
overriding future instead installs a common new answer. The design tests
whether a memory can both preserve earlier distinctions and accept
legitimate replacement. We compare a transparent frontier selector,
targeted ablations, model-written memories, simple selection rules, and
separately labeled archive-access references.

An initially favorable performance comparison does not by itself
identify a compression mechanism. A model-written memory can fail to
arrive, exceed the size cap, contain the wrong state, or be correctly
interpreted but rendered in a rejected schema. A deterministic selector
avoids some of these paths by construction. Comparing the final scores
therefore measures the implemented systems, not a pure content-selection
effect. In addition, a symbolic generator can reward arbitrary naming
conventions. Those conventions must be perturbed before original-sample
success is interpreted as robust selection.

This study consequently makes three scoped contributions. First, it
formalizes a matched-current, shared-future audit and an
all-four-future-outcomes endpoint. Second, it reconstructs the pilot's
complete result grid and gives a disjoint record-level diagnostic
analysis without deleting unsuccessful writes or redefining the primary
metric. Third, it introduces a concrete metamorphic contract for the
reference selector, implements a minimal label-equivariant repair, and
reports its negative as well as positive offline results. The repair
removes lexical identifier dependence; it does not overcome bounded
memory's uncertainty about later references.

The evidence is organized into three layers throughout: \textbf{P}, the
original paid pilot; \textbf{D}, retrospective analysis of those fixed
artifacts; and \textbf{R}, new offline repair tests on transformations
of the same source histories. Neither D nor R is an independent
confirmation of P. Both backends use the same 24 source pairs, and the
dataset is explicitly labeled development data. There is no claim of
natural-domain, long-horizon agent, or state-of-the-art algorithmic
performance.

\section{2 Related work and contribution
boundary}\label{related-work-and-contribution-boundary}

\subsection{2.1 Compression and memory
evaluation}\label{compression-and-memory-evaluation}

ACON optimizes natural-language compression guidance using trajectory
failures, and The Complexity Trap demonstrates that simple observation
masking can compete with LLM summaries in its evaluated software-agent
setting (\citeproc{ref-kang2025acon}{Kang et al. 2025};
\citeproc{ref-lindenbauer2025complexity}{Lindenbauer et al. 2025}).
These results motivate stronger baselines than an arbitrary short
summary. Our local structured writer, tail selector, and lexical ranker
do not reproduce those systems' best configurations, so outperforming
them cannot establish superiority over either published method.

TRACE evaluates compression boundaries through closed-loop continuations
from a shared environment state (\citeproc{ref-min2026trace}{Min et al.
2026}). Our pairing is different: complete historical states
intentionally differ, but their current query projections agree. A
shared future then exposes a distinction that a compressor might have
lost. This is complementary to same-state execution evaluation; a
synthetic value resolver does not replace a tool-using agent. The
Compaction Cliff further motivates protecting different kinds of
information under repeated compression
(\citeproc{ref-zerhoudi2026cliff}{Zerhoudi, Mitrovic, and Granitzer
2026}). Our two-chunk futures do not reproduce its long-running
retention setting.

MEMAUDIT is the closest neighbor for attribution. It evaluates budgeted
memory writing with explicit evidence requirements, validity state,
representation choices, and an exact package oracle
(\citeproc{ref-bhargava2026memaudit}{Bhargava and Barrento 2026}). We do
not claim novelty for retaining tombstones, separating writing from
reading, or using exact semantic checks. Our interpreter evaluates a
\emph{given} event memory; it does not optimize the best package under a
budget. The additional design studied here is the matched-current pair,
its two shared futures, and a concrete metamorphic audit of the
resulting reference selector.

\subsection{2.2 Temporal memory, reasoning, and
validation}\label{temporal-memory-reasoning-and-validation}

Zep, A-MEM, and Mem0 maintain and evolve memories using different
architectures (\citeproc{ref-rasmussen2025zep}{Rasmussen et al. 2025};
\citeproc{ref-xu2025amem}{Xu et al. 2025};
\citeproc{ref-chhikara2025mem0}{Chhikara et al. 2025}). LongMemEval and
LongMemEval-V2 already include update-sensitive capabilities; AppWorld
evaluates executable interactive agents
(\citeproc{ref-wu2024longmemeval}{Wu et al. 2024},
\citeproc{ref-wu2026longmemevalv2}{2026};
\citeproc{ref-trivedi2024appworld}{Trivedi et al. 2024}). None is
numerically ranked by this pilot. We therefore avoid the broader claim
that existing memory evaluation is uniformly static or that symbolic
retention implies real-world task success.

Reasoning-enabled summarizers are not uniformly superior in prior
dialogue-summarization experiments (\citeproc{ref-jin2025reasoning}{Jin
et al. 2025}). The present H/off writer comparison is weaker evidence
than a general reasoning study: it uses backend-specific controls, a
common output ceiling rather than matched compute, and a joint endpoint
near the floor. Its zero joint contrast is not an equivalence result.

Our relabeling checks instantiate metamorphic testing: known relations
between transformed inputs and outputs test a system even when a new
independent oracle is expensive (\citeproc{ref-chen2018metamorphic}{Chen
et al. 2018}). We claim neither to introduce metamorphic testing nor to
make transformation counts equivalent to new tasks. Finally, excluding
empty or malformed memories after treatment would change the comparison
population. The general danger of post-treatment selection is well
established (\citeproc{ref-montgomery2018conditioning}{Montgomery,
Nyhan, and Torres 2018}). We retain all planned cases and use diagnostic
cross-tabs, not a mediation estimate based on selected survivors.

\section{3 Update sufficiency and evidence
levels}\label{update-sufficiency-and-evidence-levels}

\subsection{3.1 Matched-current
histories}\label{matched-current-histories}

Let \(h\) be a history, \(q\) a query, \(u\) a future event sequence,
and \(A(h,q)\) the exact answer under public semantics. Concatenation is
\(h\oplus u\). A pair \((h_0,h_1)\) is currently equivalent for \(q\)
when

\[A(h_0,q)=A(h_1,q).\]

It is future-distinguishable relative to an update family \(\mathcal U\)
when

\[\exists u\in\mathcal U:\quad A(h_0\oplus u,q)\ne A(h_1\oplus u,q).\]

This is equality of current task outputs, not equality of hidden
historical state. A compressor writes \(m_0=C(h,q;B)\) within budget
\(B\). On the strict-memory track, subsequent state is
\(m_{t+1}=U(m_t,\delta_{t+1},q;B)\): the updater receives only the
charged preceding memory and the new chunk. A reader returns a value
from the resulting memory. Sufficiency is relative to declared queries
and updates; it is not a promise to answer arbitrary future questions
from bounded storage.

\textbf{Observation 1 (collision obstruction).} If two currently
equivalent histories have identical compressed states, downstream side
information is identical, and a shared future requires different target
answers, no deterministic updater-reader composition is correct on both.
If the compressed-state distributions are identical and the history
variant is sampled uniformly, even a randomized downstream composition
has mean correctness at most \(1/2\).

\emph{Proof.} The downstream input, or its distribution, is the same. A
deterministic output cannot equal two distinct targets. For a randomized
output, the probabilities of those mutually exclusive targets sum to at
most one. Averaging over the balanced variants gives the bound. An
archive, variant label, or other distinguishing side channel would
violate the premise. This elementary argument is a diagnostic
motivation, not a new general information-theoretic theorem.

The converse is false. Distinct memory strings can differ only in
useless wording; an adequate memory can still be misread. In the pilot,
the latest-only early states collide exactly in 12 of the 24 pairs, but
further failures occur without literal collision. Both output evaluation
and evidence inspection are needed.

\subsection{3.2 What is and is not
identified}\label{what-is-and-is-not-identified}

The implemented arm \(a\) affects selected information, generated text,
and delivery behavior. A schematic dependency is

\[a\longrightarrow (\text{content},\text{delivery})\longrightarrow m
  \longrightarrow r\longrightarrow S_q(r),\]

where \(S_q\) is the strict answer contract. Reader sampling and backend
behavior also affect \(r\). A score difference between deterministic
frontier and a model writer is consequently an end-to-end contrast
between their implemented pipelines. Holding the backend name fixed does
not equalize writer reliability, output length, or reasoning
consumption.

For event-schema memories we additionally compute \(E_q(m)\), the exact
value obtained by a separately implemented interpreter. If parsing
fails, \(E_q\) is unavailable, not automatically incorrect: a
nonconforming string can still contain an answer usable by an LLM. For
completed responses we define a diagnostic \(S_q^{\mathrm{bare}}(r)\)
that also accepts an exact bare query-value map. The original \(S_q\) is
never replaced. This yields three distinct measurements: usable strict
output, adequacy under the explicit event interpretation, and
sensitivity to one formatting convention.

These cross-classifications localize observable failure patterns but do
not identify the causal proportion of an arm gap due to content. They
neither condition the primary denominator on successful writing nor
infer what a missing response would have contained. Claims from P, D,
and R retain their distinct statuses even when all three agree.

\section{4 Pilot design and
implementation}\label{pilot-design-and-implementation}

\subsection{4.1 Histories, semantics, and
futures}\label{histories-semantics-and-futures}

The development dataset has 24 pairs: four each for rollback, expiry,
choice switch, late reference, tombstone replay, and a mixed
alias/expiry mechanism. Both models evaluate the same 48 variants. Each
prefix includes 32 distractor records. Initial event order is shuffled.
The early writer receives the target keys and public event semantics,
but the reference task constructor excludes the future text, gold
answer, variant label, and mechanism label. Thus the setting is
target-aware and future-hidden, not query-agnostic. Full wire-level
enforcement cannot be attested from the exported artifacts alone.

Events form an append-only log with unique identifiers. Identical replay
is idempotent; conflicting content with the same identifier is invalid.
A \texttt{revoke} permanently invalidates its target identifier,
including future replay. The clock is the maximum retained
\texttt{tick.now}, initially zero; expiry is inclusive. For each key,
resolution selects the greatest-revision non-revoked, unexpired
\texttt{set}, \texttt{copy}, or \texttt{choose} event. A \texttt{set}
yields a string; a \texttt{copy} resolves another key; a \texttt{choose}
resolves a selector and one of two referenced keys. Missing keys and
cycles return null. A dangling dependency of a valid selected record
returns null rather than falling back to an older version. Unique
key-revision pairs eliminate revision ties.

Each prefix is first read on the current query, then forks from the same
early memory into two futures. The \emph{reveal} future exposes the
hidden historical distinction: it revokes the latest value, advances
time, switches a selector, introduces a previously irrelevant reference,
or replays an invalidated record. The \emph{override} future installs a
new highest-revision answer common to both variants. Each future has two
chunks: the meaningful operation first and a distractor second. This is
not a long sequence of independent reversals.

Two repeats are planned for every arm and source history. For
deterministic memories, only model responses can vary; generated writers
can also vary their memories. Repeated calls, future branches, and
history variants are nested within the source pair. The configured
runner randomizes scheduling, but exact deployed order requires the
absent run ledger and lock files.

\subsection{4.2 Conditions, storage access, and
fairness}\label{conditions-storage-access-and-fairness}

All returned non-full memories are capped at \textbf{1,200 serialized
UTF-8 bytes}, including identifiers, controls, JSON syntax, and
ordering. Whole-record packers skip records that do not fit. The cap is
not 1,200 tokens. \texttt{full} is uncapped. Strict-memory methods see
preceding memory plus new events only; archive methods may inspect
accumulated raw history again after updates. An archive is an explicit
additional capability, not free bounded persistence.

The eight strict-memory conditions are \texttt{tail},
\texttt{type\_\allowbreak{}pin}, \texttt{frontier}, \texttt{frontier\_\allowbreak{}latest},
\texttt{frontier\_\allowbreak{}no\_\allowbreak{}tombstone}, \texttt{prose\_\allowbreak{}high},
\texttt{prose\_\allowbreak{}off}, and \texttt{structured\_\allowbreak{}high}. Tail packs complete
records in reverse arrival order. Type-pin prioritizes clock and
revocation events. The prose writers recursively rewrite free text;
structured H requests an events object and explicitly asks for old valid
versions, dependencies, expiry, and tombstones. The four archive
references are full history, lexical BM25 selection, dependency-closure
selection, and fresh direct H-policy prose writing from accumulated
history. Appendix A reports every arm.

Original frontier normalizes the previous records and incoming events,
filters invalid or expired value records, and collects revocations plus
the latest maximum clock. It computes the transitive key closure through
\emph{all live versions} of the query targets. Controls are ordered
first, then live events by target membership, closure membership,
revision, and identifier. Whole-record inclusion is tested against the
exact JSON byte cap. Residual capacity can preserve non-closure events.
This is a heuristic, not an optimal-memory oracle or a guarantee that
every control record fits.

Latest-only restricts live records to the largest revision per key
before closure and packing. It is not literally one scalar value:
referenced records may remain. No-tombstone first uses revocations to
filter current live records but then omits those controls from persisted
memory. Replayed invalid identifiers can therefore become active after
the deletion marker has been forgotten. These ablations manipulate
explicit persistence rules; they are not separately optimized
architectures.

\subsection{4.3 Backends, delivery, and output
scoring}\label{backends-delivery-and-output-scoring}

The exported profiles request \texttt{deepseek-flash} and
\texttt{glm-5.2} through Chat Completions. Returned names match their
respective allowlists for every nonmissing reader receipt. This checks
exported aliases, not immutable weights or router behavior. DeepSeek H
enables thinking and requests high effort; GLM H enables thinking
without an explicit effort setting. Off disables thinking for that
writer, not for its reader. Equal policy labels therefore do not
establish cross-vendor compute equivalence.

Writer and reader output allowances are 4,096 tokens, with a 600-second
configured timeout. A documented DeepSeek accounting-boundary amendment
accepts one returned receipt metered at 4,097 completion tokens. It is
still an incomplete, empty, incorrect reader response. It is not an
extra trial or a rescued success.

The memory acceptance policy is consequential. Missing infrastructure
responses propagate unknown outcomes. An over-cap returned memory is
replaced by an empty string; naturally empty output remains empty. An
incomplete response within the cap is retained and flagged. There is no
universal rollback to the last valid memory and no automatic
retry-until-success policy. Thus generated memories face delivery
failure modes that deterministic selectors avoid.

Strict reader success requires a stop-finished response of exactly
\texttt{\{"values":\ \{query\_key:\ string\_or\_null\}\}}, with
precisely the required keys and values. Invalid structure, incorrect
values, and incomplete output score zero. Infrastructure missingness
stays NA. This is the preserved operational metric, even where a less
strict consumer could use the output.

\subsection{4.4 Endpoint, missingness, and
uncertainty}\label{endpoint-missingness-and-uncertainty}

Let \(Y_{ipavb}\) be strict success for source pair \(i\), repeat \(p\),
arm \(a\), variant \(v\), and branch \(b\). The configured joint
endpoint is

\[J_{ipa}=\prod_{v\in\{0,1\}}\prod_{b\in\{\mathrm{reveal},\mathrm{override}\}}Y_{ipavb}.\]

Current performance is reported separately and does not filter
eligibility. An arm has 48 pair-repeat blocks, but only 24 source-pair
clusters. We average repeats within each pair before averaging pairs.
This endpoint is intentionally demanding: success requires retaining
history-sensitive information \emph{and} accepting a replacing update
for both variants.

For an incomplete block, any observed zero identifies \(J=0\).
Otherwise, a block with an unresolved constituent has bounds \([0,1]\).
Averaging block lower and upper values yields \([L_a,U_a]\). The
frontier-minus-baseline contrast is bounded by

\[\Delta\in[L_{\mathrm{frontier}}-U_{\mathrm{baseline}},\;
U_{\mathrm{frontier}}-L_{\mathrm{baseline}}].\]

These are finite-sample identification bounds over unobserved outcomes,
not confidence intervals or tests of a population null. All planned
denominators remain fixed, without a missing-at-random assumption.

The archived analysis also resamples pairs within each of the six fixed
families, retaining their repeats and branches: 2,000 draws with seed
20260911. It reports the 2.5th percentile of lower endpoints and 97.5th
percentile of upper endpoints, following the resampling principle of the
bootstrap (\citeproc{ref-efron1979bootstrap}{Efron 1979}). With only
four pairs per family, this is an approximate conditional endpoint
envelope, not a calibrated simultaneous confidence set for unfamiliar
mechanisms.

The supplied configuration names frontier versus structured and prose
off versus H across three planned model slots; only two slots have
results. The unused model is not silently removed to obtain a smaller
multiplicity family. Holm correction does not repair invalid underlying
tests (\citeproc{ref-holm1979simple}{Holm 1979}). No confirmatory
frontier p-value is reported. Configuration intent is not evidence of
externally timestamped preregistration.

\section{5 Results: paid pilot and retrospective
attribution}\label{results-paid-pilot-and-retrospective-attribution}

\subsection{5.1 Complete exported grid, bounded
provenance}\label{complete-exported-grid-bounded-provenance}

Both backends contain all 3,456 planned outcome records, with no
duplicate or extra identifiers. DeepSeek has one NA and GLM three.
Independent rescoring reproduces every original score and failure class.
Dataset and gold hashes match the reference kit, and a separately
implemented interpreter verifies every supplied gold answer.

The exported evidence supports \textbf{5,181 done operations on DeepSeek
and 5,180 on GLM}, or \textbf{10,361 done}, not 10,362. Including four
missing operations gives 10,365 recorded operations below the 10,368
ceiling. Three planned downstream calls were not dispatched after
upstream missing writers. The boundary-settlement receipt is already
present in the outcomes and is not counted twice. Complete exported
grids establish completion of the recorded analysis surface, not
live-process exit or invoice reconciliation.

We additionally replay the eight deterministic reference conditions from
source histories, reconstructing 1,152 distinct branch memories and
comparing every corresponding repeat/backend export. \textbf{All 4,608
exported memories match byte-for-byte.} This strengthens functional
reproducibility for those states. It does not attest the complete
deployed source, model-written requests, or absence of external side
channels.

\subsection{5.2 Current success and update success
diverge}\label{current-success-and-update-success-diverge}

Table 1. Selected original strict results (P), with fixed denominators.
DS = DeepSeek. Current and reveal columns have 96 planned reads; joint
columns have 48 pair-repeat blocks within 24 pairs. An asterisk
indicates one unresolved case or block, not an added success. Full
archive is not storage-equivalent to strict memory.

\begin{longtable}[]{@{}
  >{\raggedright\arraybackslash}p{(\columnwidth - 12\tabcolsep) * \real{0.1429}}
  >{\raggedright\arraybackslash}p{(\columnwidth - 12\tabcolsep) * \real{0.1429}}
  >{\raggedright\arraybackslash}p{(\columnwidth - 12\tabcolsep) * \real{0.1429}}
  >{\raggedright\arraybackslash}p{(\columnwidth - 12\tabcolsep) * \real{0.1429}}
  >{\raggedright\arraybackslash}p{(\columnwidth - 12\tabcolsep) * \real{0.1429}}
  >{\raggedright\arraybackslash}p{(\columnwidth - 12\tabcolsep) * \real{0.1429}}
  >{\raggedright\arraybackslash}p{(\columnwidth - 12\tabcolsep) * \real{0.1429}}@{}}
\toprule\noalign{}
\begin{minipage}[b]{\linewidth}\raggedright
Condition
\end{minipage} & \begin{minipage}[b]{\linewidth}\raggedright
DS current
\end{minipage} & \begin{minipage}[b]{\linewidth}\raggedright
DS reveal
\end{minipage} & \begin{minipage}[b]{\linewidth}\raggedright
DS joint
\end{minipage} & \begin{minipage}[b]{\linewidth}\raggedright
GLM current
\end{minipage} & \begin{minipage}[b]{\linewidth}\raggedright
GLM reveal
\end{minipage} & \begin{minipage}[b]{\linewidth}\raggedright
GLM joint
\end{minipage} \\
\midrule\noalign{}
\endhead
\bottomrule\noalign{}
\endlastfoot
Frontier & 96/96 & 96/96 & 45/48 & 78/96 & 82/96 & 28/48 \\
Latest-only & 96/96 & 32/96 & 15/48 & 84/96 & 23/96 & 5/48 \\
No tombstones & 95/96 & 79/96 & 38/48 & 82/96 & 65/96 & 20/48 \\
Structured H & 76/96 & 62/96* & 19/48* & 64/96 & 56/96 & 13/48* \\
Full archive & 96/96 & 95/96 & 46/48 & 83/96 & 75/96 & 23/48 \\
\end{longtable}

The clearest dissociation is DeepSeek frontier versus latest-only. Both
score 96/96 on the current branch; reveal falls from 96/96 to 32/96 when
older live records are removed. On GLM, latest-only has higher current
strict accuracy, 84/96 versus 78/96, but lower reveal accuracy, 23/96
versus 82/96. Current-answer performance alone can therefore prefer a
much less update-capable memory on these histories. This is not a claim
that static quality and updateability are statistically independent.

\begin{figure}
\centering
\includegraphics[width=0.95\textwidth,height=\textheight]{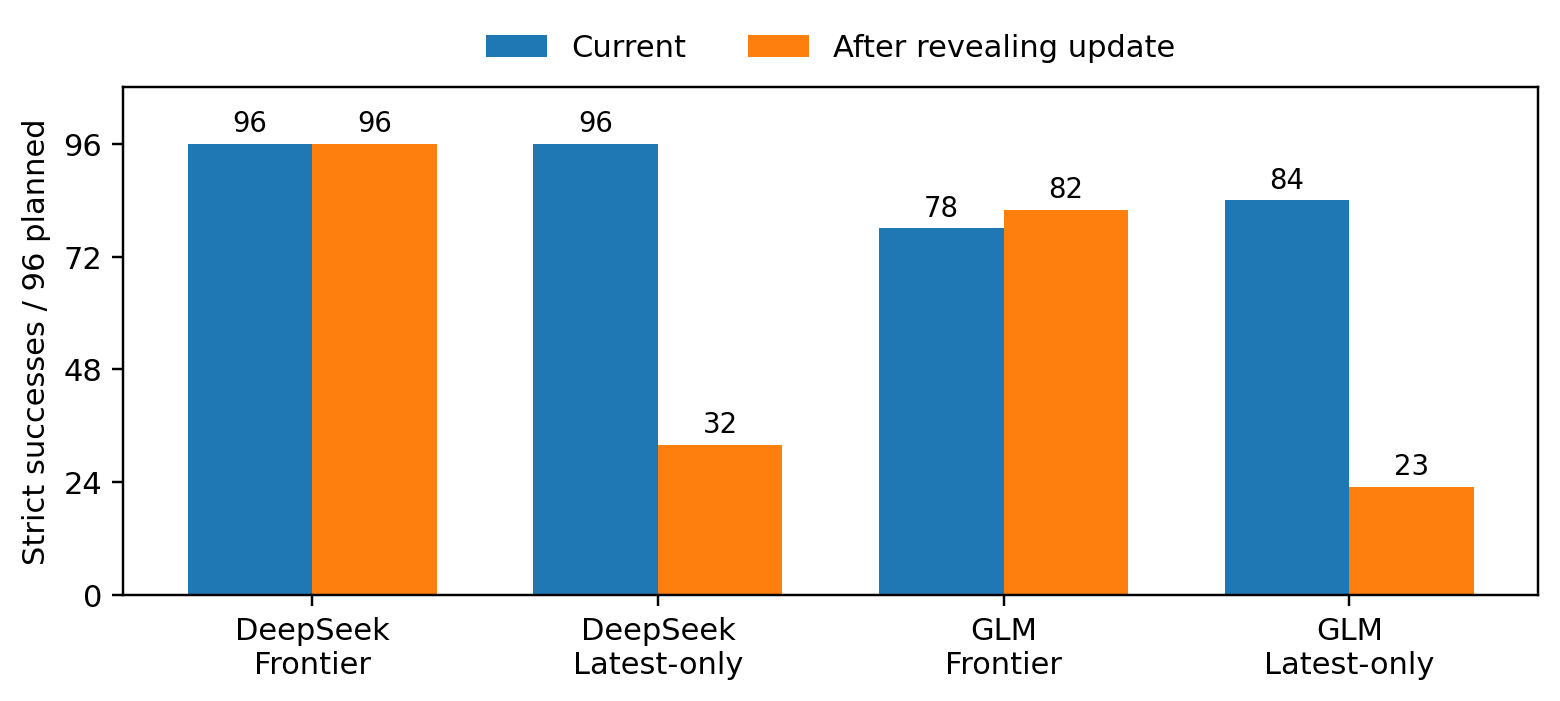}
\caption{Current and reveal strict successes on the original paid pilot.
Shared source pairs and repeated reads are not independent tasks.
Numerical values are also in Table 1.}
\end{figure}

Table 2. Frontier minus structured H: original fixed-sample
identification bounds and approximate conditional bootstrap envelopes.
Joint and reveal are different endpoints; none of these bounds
establishes external-domain performance.

\begin{longtable}[]{@{}
  >{\raggedright\arraybackslash}p{(\columnwidth - 6\tabcolsep) * \real{0.2500}}
  >{\raggedright\arraybackslash}p{(\columnwidth - 6\tabcolsep) * \real{0.2500}}
  >{\raggedright\arraybackslash}p{(\columnwidth - 6\tabcolsep) * \real{0.2500}}
  >{\raggedright\arraybackslash}p{(\columnwidth - 6\tabcolsep) * \real{0.2500}}@{}}
\toprule\noalign{}
\begin{minipage}[b]{\linewidth}\raggedright
Endpoint
\end{minipage} & \begin{minipage}[b]{\linewidth}\raggedright
Backend
\end{minipage} & \begin{minipage}[b]{\linewidth}\raggedright
Identification interval
\end{minipage} & \begin{minipage}[b]{\linewidth}\raggedright
Approx. bootstrap envelope
\end{minipage} \\
\midrule\noalign{}
\endhead
\bottomrule\noalign{}
\endlastfoot
Joint & DeepSeek & {[}0.520833, 0.541667{]} & {[}0.395833,
0.646354{]} \\
Reveal & DeepSeek & {[}0.343750, 0.354167{]} & {[}0.260417,
0.447917{]} \\
Joint & GLM & {[}0.291667, 0.312500{]} & {[}0.104167, 0.479167{]} \\
Reveal & GLM & {[}0.270833, 0.270833{]} & {[}0.156250, 0.385417{]} \\
\end{longtable}

The prominent joint intervals, {[}0.520833, 0.541667{]} and {[}0.291667,
0.312500{]}, must not be presented as reveal-only gaps. The reveal
difference is 34.38--35.42 percentage points on DeepSeek and 27.08
points on GLM. Frontier's strict joint score is 45/48 on DeepSeek, below
full history's 46/48; on GLM it is 28/48 versus 23/48. Neither pattern
establishes a general advantage over complete raw-history access.

Pair-level heterogeneity is retained. Against structured H, DeepSeek has
16 definitely positive pair-mean joint contrasts, seven ties, and one
negative; GLM has 12 positive, eight ties, three negative, and one
unresolved contrast. A leave-one-family-out descriptive audit keeps the
joint lower endpoint positive for every omission: its minimum is 0.45 on
DeepSeek and 0.20 on GLM. These checks show the aggregate does not
depend on only one family, but remain reuse of development data, not
replication on omitted-family training or unseen-family evaluation.

\subsection{5.3 A joint content-and-response
diagnostic}\label{a-joint-content-and-response-diagnostic}

Table 3 partitions all 96 planned reveal records for each selected
arm/backend, without changing their strict scores. ``Unparsed'' means
that the event-schema interpreter cannot evaluate the retained memory.
``Wrong log'' means a valid log resolves to the wrong value. The last
three columns all have a semantically correct event log but differ in
reader output. These categories are diagnostic observations, not
estimated causal components.

Table 3. Retrospective reveal cross-classification (D). Columns are
mutually exclusive and sum to 96. Wrapper-only failures satisfy the
exact bare-map sensitivity rule. Wrong-log cells in these selected rows
have no strict successes; the implementation nevertheless preserves that
possible discordance as a separate category.

\begin{longtable}[]{@{}
  >{\raggedright\arraybackslash}p{(\columnwidth - 14\tabcolsep) * \real{0.1250}}
  >{\raggedright\arraybackslash}p{(\columnwidth - 14\tabcolsep) * \real{0.1250}}
  >{\raggedright\arraybackslash}p{(\columnwidth - 14\tabcolsep) * \real{0.1250}}
  >{\raggedright\arraybackslash}p{(\columnwidth - 14\tabcolsep) * \real{0.1250}}
  >{\raggedright\arraybackslash}p{(\columnwidth - 14\tabcolsep) * \real{0.1250}}
  >{\raggedright\arraybackslash}p{(\columnwidth - 14\tabcolsep) * \real{0.1250}}
  >{\raggedright\arraybackslash}p{(\columnwidth - 14\tabcolsep) * \real{0.1250}}
  >{\raggedright\arraybackslash}p{(\columnwidth - 14\tabcolsep) * \real{0.1250}}@{}}
\toprule\noalign{}
\begin{minipage}[b]{\linewidth}\raggedright
Model
\end{minipage} & \begin{minipage}[b]{\linewidth}\raggedright
Condition
\end{minipage} & \begin{minipage}[b]{\linewidth}\raggedright
NA
\end{minipage} & \begin{minipage}[b]{\linewidth}\raggedright
Unparsed
\end{minipage} & \begin{minipage}[b]{\linewidth}\raggedright
Wrong log
\end{minipage} & \begin{minipage}[b]{\linewidth}\raggedright
Correct + pass
\end{minipage} & \begin{minipage}[b]{\linewidth}\raggedright
Wrapper only
\end{minipage} & \begin{minipage}[b]{\linewidth}\raggedright
Other reader
\end{minipage} \\
\midrule\noalign{}
\endhead
\bottomrule\noalign{}
\endlastfoot
DS & Frontier & 0 & 0 & 0 & 96 & 0 & 0 \\
DS & Latest-only & 0 & 0 & 64 & 32 & 0 & 0 \\
DS & No tombstones & 0 & 0 & 16 & 79 & 1 & 0 \\
DS & Structured H & 1 & 7 & 26 & 62 & 0 & 0 \\
GLM & Frontier & 0 & 0 & 0 & 82 & 14 & 0 \\
GLM & Latest-only & 0 & 0 & 64 & 23 & 8 & 1 \\
GLM & No tombstones & 0 & 0 & 16 & 65 & 14 & 1 \\
GLM & Structured H & 0 & 16 & 25 & 49 & 6 & 0 \\
\end{longtable}

This partition sharpens attribution in two directions. First, structured
H has 26 valid but wrong reveal logs on DeepSeek and 25 on GLM. Its
problems are not confined to missing or syntactically malformed outputs.
Second, all 96 frontier reveal memories are adequate under the exact
interpreter on both backends. The 14 strict GLM frontier failures are
therefore not cases of missing event evidence: all contain the correct
bare value map but omit the required wrapper.

The event parser is not a universal semantic judge. Of GLM's 16 unparsed
structured reveal memories, seven still yield strict-correct reader
answers. Those strings must not be labeled ``information destroyed''
merely because they are not valid event envelopes. Similarly, a wrong
retained state and an accidentally correct answer would remain a
discordant observation, not be rewritten to fit a neat failure
narrative. All 6,912 result rows are assigned exactly once by the audit.

The alternative reader rule \(S_q^{\mathrm{bare}}\) accepts only an
additional stop-finished JSON object with exactly the query keys and
correct string/null values. It does not remove code fences, infer keys,
complete truncated text, or invoke an LLM judge. Applied uniformly to
all arms, it changes GLM frontier reveal from 82/96 to 96/96 and
structured H from 56/96 to 63/96; DeepSeek stays 96/96 and 62 successes
with one NA. The primary operational scores are untouched. Consequently,
the smaller strict GLM gap cannot be interpreted directly as weaker
content retention or value inference.

Under this sensitivity frontier joint success becomes 48/48 on DeepSeek
and 45/48 on GLM. The corresponding structured joint counts are 19 and
17 known successes, each with one unresolved block. These are post-hoc
diagnostic results, not replacement primary effects. Their purpose is to
identify answer-contract dependence for a future frozen evaluation.

\subsection{5.4 Localized ablations and delivery
failures}\label{localized-ablations-and-delivery-failures}

The event interpreter resolves all original frontier, full, and
closure-archive memories correctly. Latest-only resolves 32/96 reveal
cases; no-tombstone resolves 80/96. The latter's 16 failures are exactly
the tombstone-replay mechanism: forgotten revocations allow invalid
intermediate records to reappear. Actual strict reveal success in that
stratum is 0/16 under no-tombstone on both models, versus 16/16 and
15/16 for frontier. This supports the specified persistence mechanism
without requiring a language-model judge.

It would nevertheless be inaccurate to say the entire no-tombstone arm
collapses. Its overall reveal scores are 79/96 and 65/96. Likewise,
latest-only remains effective on choice-switch and original
late-reference instances. Each family has four independent source pairs;
sixteen repeated/variant outcomes are not sixteen independent
demonstrations. Appendix A gives the complete mechanism table.

Delivery remains a substantial competing explanation for comparisons
with generated writers. DeepSeek prose H has 115 raw-empty writes among
480 completed operations; direct H has 157 among 288. GLM has 85/480 and
47/288, respectively. Structured writers have 28 and 19 raw-empty
returns. Empty and incomplete columns overlap and are not summed
(Appendix B). The acceptance policy can also empty an over-cap result.
Thus the pilot has not structurally eliminated empty-summary failures.

The H/off joint contrast is zero because DeepSeek has one passing joint
block in each prose condition and GLM has none. Their reveal counts are
not equal: 15 versus 14 on DeepSeek and 11 versus three on GLM. A
floor-limited joint endpoint, equal output allowances, and different
visible-output delivery cannot establish that reasoning policy has no
effect. Neither discarding empty runs nor giving only one arm unlimited
retries would repair the original comparison.

\section{6 Metamorphic audit and a bounded reference
repair}\label{metamorphic-audit-and-a-bounded-reference-repair}

\subsection{6.1 Identifier relabeling exposes an incidental
shortcut}\label{identifier-relabeling-exposes-an-incidental-shortcut}

Original frontier uses identifier lexicographic order to resolve
priority ties. In a post-hoc symbolic stress test, we replace every
event identifier and key name with a unique random ASCII string of the
same length, consistently rewriting references and query keys. Values,
revisions, time, dependency topology, event order, and the 1,200-byte
budget are unchanged. Full-history answers are independently verified
after every transformation. There are 40 fixed renamings per source
pair; the transformed instances are not new independent samples.

Original frontier late-reference adequacy drops from 8/8 original
variants to 94/320 renamed variants. Each of the other five mechanisms
remains 320/320. A previously irrelevant record is retained only when
residual capacity and the lexical tie-break favor it. Closure archive
continues to resolve all late-reference answers because it can access
raw history after the new reference appears; that result relies on a
different storage capability.

This is evidence of sensitivity to irrelevant naming, not proof that all
earlier-version retention is ineffective. It also makes
unchanged-generator expansion insufficient: more pairs with the same
advantageous naming convention would not test the exposed failure.

\subsection{6.2 A label-equivariant selector, not an improved paid
result}\label{a-label-equivariant-selector-not-an-improved-paid-result}

We implement \textbf{frontier-alpha} after the pilot. It keeps the
original control selection, all-version dependency closure, whole-record
byte accounting, and strict update interface. The sole priority change
removes lexical identifiers from both control and value tie-breaks.
Equal-priority items preserve the order in which they occur in the
current normalized input. That order is carried by the budgeted event
array itself; there is no uncharged arrival-rank table or raw archive.

Let \(\pi\) consistently rename the event identifiers and key namespace
without changing values, revisions, timestamps, equality relations,
record order, or the serialized size of any record. For the canonical
JSON encoding, the checked family uses same-length ASCII labels. The
relevant contract is

\[C_\alpha(\pi(h),\pi(q);B)=\pi\bigl(C_\alpha(h,q;B)\bigr),\]

where the right-hand side includes decoding and canonically serializing
the renamed event memory. The analogous property holds for each update
when its previous memory and incoming chunk are renamed consistently.

\textbf{Proposition 2 (label equivariance).} Under those premises,
frontier-alpha satisfies the contract at initial compression and at
every strict-memory update, including when records are skipped because
of the byte cap.

\emph{Proof.} Normalization, identifier equality, revocation membership,
live-version selection, target membership, and dependency closure all
commute with a bijective renaming. Alpha's priorities contain no lexical
label comparison. Equal priorities preserve corresponding input
positions, so the ordered candidate lists correspond. Every candidate
has the same encoded length before and after renaming; induction over
the packing loop therefore gives identical inclusion decisions. The
resulting event arrays correspond under \(\pi\). Decoding the preceding
array and appending the correspondingly renamed next chunk preserves the
premises, completing induction over updates. A cap below the 13-byte
empty event envelope is rejected on both sides, rather than silently
returning an invalid memory.

The premise concerns \textbf{serialized length}, not merely the number
of Unicode characters. Escaping or different UTF-8 lengths can change
capacity and invalidate a naive relabeling test. The proposition says
nothing about permutation invariance, unknown future keys, optimal
packing, or LLM reader performance.

\subsection{6.3 The repair removes label dependence but not future
uncertainty}\label{the-repair-removes-label-dependence-but-not-future-uncertainty}

We run both selectors on identity inputs, 40 same-length renamings, and
40 prefix permutations per pair. The permutations are shared by the two
variants, preserve full-history answers under the event semantics, and
do not alter the future. All intermediate memories are checked against
the byte cap. The audit executes 13,440 stage-level alpha equivariance
comparisons, with zero violations. These checks reuse source histories
and repeated early states; their count is not a statistical sample size.

\Needspace{12\baselineskip}

Table 4. Late-reference exact retained-state adequacy after reveal (R),
at 1,200 bytes. This is an offline interpreter result, not a model
score. Each row derives from the same four source pairs; 320 is eight
variants times 40 transformations.

\begin{longtable}[]{@{}lll@{}}
\toprule\noalign{}
Offline input condition & Original frontier & Frontier-alpha \\
\midrule\noalign{}
\endhead
\bottomrule\noalign{}
\endlastfoot
Original instances & 8/8 & 2/8 \\
Same-length renaming & 94/320 & 80/320 \\
Prefix permutations & 320/320 & 78/320 \\
\end{longtable}

On the original late-reference instances, alpha retains only 2/8 needed
values. Under 40 renamings it retains exactly the corresponding 80/320,
as equivariance requires. Under prefix permutations it retains 78/320.
Original frontier remains 320/320 under these permutations because its
lexical priority is unchanged. Each other family is 8/8 at identity and
320/320 under each transformation for both selectors. Thus protecting
versions and tombstones survives these particular perturbations, whereas
unknown late relevance remains unresolved.

\begin{figure}
\centering
\includegraphics[width=0.92\textwidth,height=\textheight]{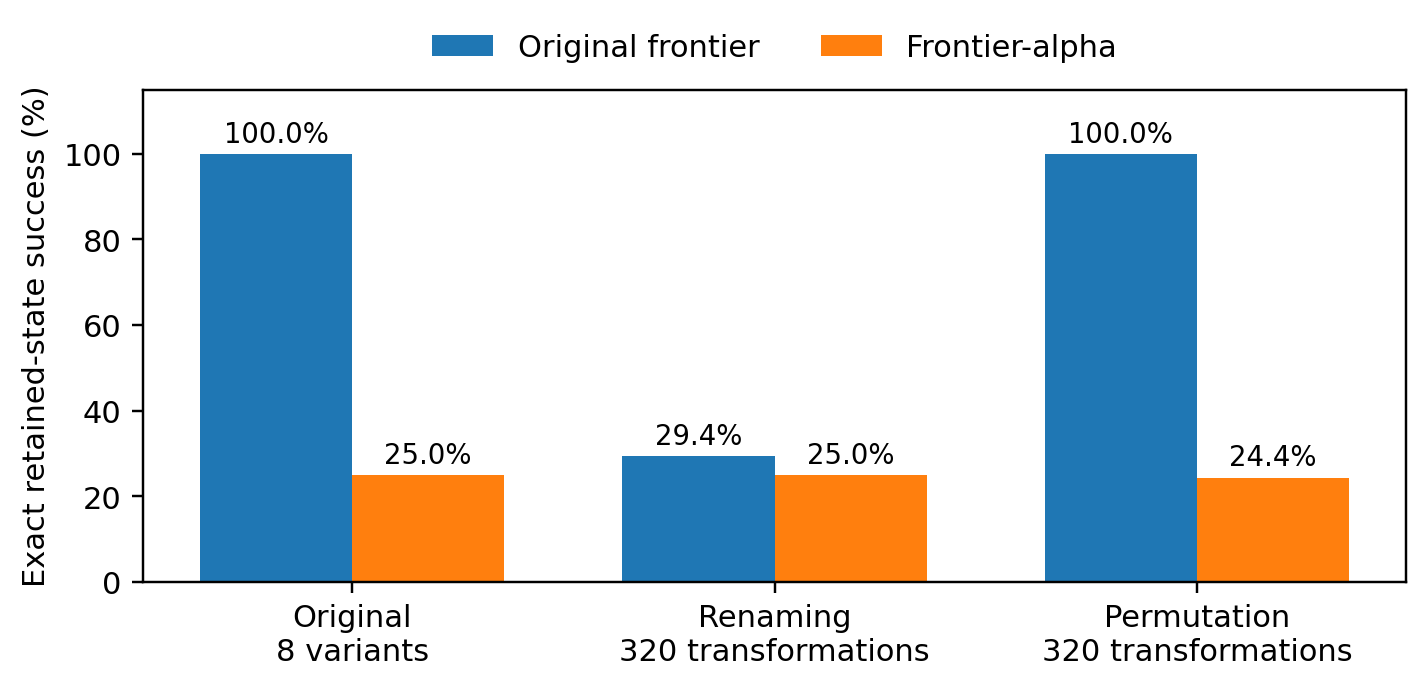}
\caption{Late-reference retained-state adequacy. The repaired selector
removes naming dependence without improving adequacy. Transformation
counts reuse four source pairs and do not supply independent trials or
error bars.}
\end{figure}

This negative repair result is scientifically useful. It separates an
avoidable nuisance dependence from a substantive capacity/relevance
problem. An invariant algorithm can be consistently inadequate; a
high-scoring algorithm can exploit an incidental generator convention.
Neither score alone validates a general compression mechanism.

A six-cap offline sweep further illustrates the boundary. On the
original eight late-reference variants, alpha resolves 2/8 at 600, 900,
1,200, and 1,800 bytes, 4/8 at 2,400, and 8/8 at 4,800. These values are
descriptive, not a tuned cap recommendation. Larger capacity can
preserve more unrelated evidence, but does not make the fixed-budget
selector universally sufficient.

\textbf{Observation 3 (arbitrary late-query limit).} Suppose a history
contains \(N\) independently chosen values, each from an alphabet of
size \(d\), and a future can ask for any indexed value. With no side
channel, a deterministic finite-state memory that answers every such
query exactly must distinguish all \(d^N\) assignments. If its
representation has at most \(2^b\) states, then \(b\ge N\log_2 d\).

\emph{Proof.} If fewer states encode all assignments, two different
assignments collide. They differ at an index; querying that index
requires different outputs from identical memory and side information.
This contradicts exact recovery. This elementary counting argument is
not a tight bound for our JSON packer, nor an impossibility claim about
likely queries or archive-assisted systems. It makes the declared future
family and storage budget explicit.

\section{7 Reproducibility and validation
scope}\label{reproducibility-and-validation-scope}

The artifact distributes sanitized outcome exports, source-data hashes,
matching histories and gold values, the reference selector code, the
alpha repair, all analysis programs, generated tables, and tests. No API
call is needed to reproduce D or R. The paid data are kept unchanged.
New scripts produce separate diagnostic and repair files rather than
overwriting the original outcome labels.

The strongest executable checks are functional rather than historical:
exact gold replay, all original scores, the complete grid, distinct
operation accounting, 4,608 byte-identical deterministic memories, and
alpha's checked transformation contract. None reconstructs missing
writer wire payloads or a server-side model snapshot. The result archive
lacks the full live SQLite ledgers, run-lock manifests, some raw writer
texts, and complete deployed-source attestation. Returned aliases are
not immutable weights; a model-name amendment must not be interpreted as
proof of unchanged underlying computation.

Known deduplicated receipts report 4,597,572 input and 4,196,900
completion tokens for DeepSeek, and 4,412,518 input and 5,623,634
completion tokens for GLM. Unknown usage for missing operations is not
included. The artifacts do not establish complete currency cost, cache
billing, or equal compute across vendors. We make no latency or
monetary-efficiency superiority claim.

The evidence supports three bounded conclusions. On these development
histories, current-answer success can conceal future-update failure.
Under the specified semantics, forgetting versions and revocations
causes localized content failures. Finally, response formatting and
incidental label order materially affect apparent performance. The
evidence does \textbf{not} establish a robust general compressor,
external-benchmark superiority, an equivalent H/off writing policy, a
repaired model result, or a pure semantic effect of frontier versus
equally reliable structured writing.

The repair addresses a reproducible implementation dependency; the
cross-tabs correct attribution language; fixed denominators and evidence
levels correct validation claims. They do not manufacture missing
external evidence. A complete manuscript can report an honest
exploratory mechanism study while still requiring additional work for
stronger claims.

\section{8 A prospective confirmation
contract}\label{a-prospective-confirmation-contract}

A confirmatory study should not simply increase the unchanged generator
to 192 pairs. Its implementation and measurement contracts must first be
repaired on development material, and all resulting changes require new
experiment identities. The following design is proposed, \textbf{not
executed} in this paper.

First, separate representation selection from delivery. Compare
deterministic alpha, a structured writer, a simple whole-record
baseline, and targeted ablations under explicit storage parity. A common
deterministic delivery layer should validate the envelope, enforce the
byte cap, and record whether the candidate or a bounded fallback was
committed. The original reject-to-empty policy and the validated-commit
policy should be crossed with applicable methods, rather than improving
only the favored arm. A fallback must use only charged prior memory and
the new chunk, never an uncharged archive or future labels. Extra
requests and fallback frequency are outcomes and costs, not grounds to
discard a trial.

Second, use a fixed semantic interface without erasing operational
failures. Before running confirmation, declare strict response validity
as one endpoint and exact value accuracy as a separate secondary
endpoint; specify the accepted schemas without reference to new
outcomes. Persist raw output and finish status. A length-finished or
transport-missing response is not silently repaired into a completed
success. Paired source histories, complete outcome grids, missingness
bounds, and source-level clustering remain fixed.

Third, distinguish \textbf{within-generator confirmation} from
\textbf{external validation}. New seeds alone supply new instances of an
old mechanism. Independently authored mechanisms and source-separated
natural or executable tasks are needed for transfer claims. Freeze task
sources, retrieval permissions, query visibility, budgets,
randomization, and stopping criteria before exposing outcomes. Renaming
and order tests should be included prospectively, not selected afterward
because a candidate passes. No result from the revised development audit
can be relabeled as held-out evidence.

Finally, freeze the primary contrast, model identities or documented
alias limits, and the multiplicity family before choosing a confirmatory
sample size. Estimate cost and precision on development data, not on
sealed-test success. LongMemEval-V2 and AppWorld offer complementary
evaluation settings, but adapting them is a separate study, not a result
of this pilot (\citeproc{ref-wu2026longmemevalv2}{Wu et al. 2026};
\citeproc{ref-trivedi2024appworld}{Trivedi et al. 2024}). The supplied
confirmation contract has execution disabled and does not initiate paid
work.

\section{9 Limitations}\label{limitations}

The dataset has only four pairs per mechanism, and mechanisms share one
synthetic grammar. Its selector understands that grammar directly. A
successful interpreter reconstruction therefore does not establish
useful compression of natural conversations, repositories, or partially
observed environments. Two update chunks are not a long-horizon stress
test. Repeated calls provide operational observations but do not create
independent task diversity.

Our structured baseline is informed about relevant event types, but is
neither delivery-matched to deterministic selection nor a faithful
reproduction of the strongest published systems. The diagnostic tables
cannot recover its counterfactual performance under repaired writing.
The event interpreter has deliberately narrow coverage; seven GLM
structured successes from unparsed memory demonstrate why parser failure
must not be equated with semantic loss. Format sensitivity and selector
repair were developed after seeing the pilot.

Alpha establishes label equivariance only under its stated
transformation premises. It remains sensitive to record order and
unknown late relevance. Its tests reuse the original pairs, use an exact
symbolic reader, and have no live backend evaluation. A zero violation
count over transformations is not a bound on real-task failure
probability, and a proof about labels is not a proof of update
sufficiency.

Statistical inference is likewise scoped. Identification bounds account
for unresolved outcomes within the fixed sample. The small-stratum
bootstrap is an approximate development-distribution diagnostic. Neither
makes model backends, repeats, variants, or transformations into
independent samples. The selected methods, incomplete selection history,
missing deployed provenance, and unrun planned third model preclude a
retrospective confirmatory significance claim.

\section{10 Conclusion}\label{conclusion}

The CCA v4 pilot shows why current correctness is not enough to evaluate
compressed memory under updates. A matched-current pair can require
different answers to the same future, and targeted deletion of older
versions or tombstones destroys those distinctions in the mechanisms
that need them. Both backends favor original frontier over the
configured structured writer on a demanding joint endpoint, but that
comparison includes delivery and answer-contract effects.

A fuller audit changes the scientific conclusion rather than merely
adding caveats. It distinguishes 26 and 25 valid-but-wrong structured
reveal logs from GLM's 14 correct-value frontier formatting failures. It
also exposes identifier-order dependence and provides a
label-equivariant repair whose late-reference adequacy is lower on the
original instances. Correcting a shortcut is not evidence of a stronger
compressor. The defensible outcome is a reproducible, bounded
methodology for evaluating update sufficiency and diagnosing its failure
modes, with an explicit path to stronger prospective validation.

\section*{Ethics, data, and AI
assistance}\label{ethics-data-and-ai-assistance}
\addcontentsline{toc}{section}{Ethics, data, and AI assistance}

The analyzed tasks are synthetic event records, not observations of
human participants. Distributed derivative receipts omit provider
request identifiers and system fingerprints while retaining
analysis-relevant content and provenance hashes. Applying persistent
memory to people would require consent, access control, and deletion
safeguards beyond this study; remembering invalidation does not
authorize retaining personal data.

AI systems assisted research-design discussion, analysis-code development,
artifact auditing, diagnostic analysis, the reference repair, and manuscript
preparation. This assistance extended beyond language editing. No AI system
is an author. Automated tests and
multiple author-side revision passes are not independent peer review.
The human author is responsible for checking the final manuscript, the
provenance of the paid runs, release permissions, and any venue-specific
disclosures. No unsupported funding or conflict-of-interest declaration
is inferred from the supplied materials.

\section*{References}\label{references}
\addcontentsline{toc}{section}{References}

\phantomsection\label{refs}

\newpage

\section{Appendix A. Complete paid results and mechanism
localization}\label{appendix-a.-complete-paid-results-and-mechanism-localization}

Table A1. All original strict branch results. Every cell has 96 planned
outcomes. Asterisk: one NA. C = current; R = reveal; O = override.
Archive-access conditions must not be treated as storage-equivalent to
strict bounded memory.

\begin{longtable}[]{@{}lllllll@{}}
\toprule\noalign{}
Condition & DS C & DS R & DS O & GLM C & GLM R & GLM O \\
\midrule\noalign{}
\endhead
\bottomrule\noalign{}
\endlastfoot
Frontier & 96/96 & 96/96 & 93/96 & 78/96 & 82/96 & 85/96 \\
Latest-only & 96/96 & 32/96 & 95/96 & 84/96 & 23/96 & 87/96 \\
No tombstones & 95/96 & 79/96 & 95/96 & 82/96 & 65/96 & 84/96 \\
Structured H & 76/96 & 62/96* & 89/96 & 64/96 & 56/96 & 80/96* \\
Prose H & 46/96 & 15/96 & 60/96 & 20/96* & 11/96 & 30/96 \\
Prose off & 73/96 & 14/96 & 75/96 & 22/96 & 3/96 & 31/96* \\
Tail & 24/96 & 4/96 & 0/96 & 16/96 & 4/96 & 0/96 \\
Type-pin & 24/96 & 16/96 & 0/96 & 19/96 & 10/96 & 0/96 \\
Full archive & 96/96 & 95/96 & 95/96 & 83/96 & 75/96 & 86/96 \\
Closure archive & 93/96 & 96/96 & 94/96 & 80/96 & 76/96 & 79/96 \\
BM25 archive & 79/96 & 35/96 & 96/96 & 60/96 & 29/96 & 83/96 \\
Direct H archive & 42/96 & 42/96 & 39/96 & 26/96 & 26/96 & 28/96 \\
\end{longtable}

Table A2. All original joint scores. Every cell has 48 pair-repeat
blocks within 24 source pairs. Asterisk: one unresolved block. Current
outcomes are not part of this endpoint.

\begin{longtable}[]{@{}lll@{}}
\toprule\noalign{}
Condition & DeepSeek joint & GLM joint \\
\midrule\noalign{}
\endhead
\bottomrule\noalign{}
\endlastfoot
Frontier & 45/48 & 28/48 \\
Latest-only & 15/48 & 5/48 \\
No tombstones & 38/48 & 20/48 \\
Structured H & 19/48* & 13/48* \\
Prose H & 1/48 & 0/48 \\
Prose off & 1/48 & 0/48 \\
Tail & 0/48 & 0/48 \\
Type-pin & 0/48 & 0/48 \\
Full archive & 46/48 & 23/48 \\
Closure archive & 46/48 & 24/48 \\
BM25 archive & 17/48 & 8/48 \\
Direct H archive & 4/48 & 0/48 \\
\end{longtable}

Table A3. Strict reveal counts by mechanism. F = frontier; L =
latest-only; NT = no tombstones. Every cell has 16 planned evaluations
but only four independent pairs. This is mechanism localization, not six
natural-domain replications.

\begin{longtable}[]{@{}lllllll@{}}
\toprule\noalign{}
Mechanism & DS F & DS L & DS NT & GLM F & GLM L & GLM NT \\
\midrule\noalign{}
\endhead
\bottomrule\noalign{}
\endlastfoot
Rollback & 16/16 & 0/16 & 16/16 & 12/16 & 0/16 & 13/16 \\
Expiry & 16/16 & 0/16 & 16/16 & 15/16 & 0/16 & 14/16 \\
Choice switch & 16/16 & 16/16 & 15/16 & 13/16 & 11/16 & 12/16 \\
Late reference & 16/16 & 16/16 & 16/16 & 15/16 & 12/16 & 13/16 \\
Tombstone replay & 16/16 & 0/16 & 0/16 & 15/16 & 0/16 & 0/16 \\
Mixed & 16/16 & 0/16 & 16/16 & 12/16 & 0/16 & 13/16 \\
\end{longtable}

The latest-only condition's twelve exact early-memory collisions do not
account for all its failures: losing a dependency can invalidate an
answer even when remaining irrelevant records distinguish the strings.
Conversely, full history is an information-access reference, not a
perfect LLM reader. Its occasional strict errors are consistent with
adequate evidence and imperfect inference or output rendering.

\section{Appendix B. Delivery, missingness, and cost
accounting}\label{appendix-b.-delivery-missingness-and-cost-accounting}

Table B1. Distinct logical writer operations. The early write shared by
current, reveal, and override is counted once. Raw-empty, over-cap, and
incomplete columns can overlap. NA is not a zero-cost or
semantic-failure label.

\begin{longtable}[]{@{}lllllll@{}}
\toprule\noalign{}
Backend & Writer & Done & NA & Empty & Over cap & Incomplete \\
\midrule\noalign{}
\endhead
\bottomrule\noalign{}
\endlastfoot
DS & Prose H & 480 & 0 & 115 & 0 & 121 \\
DS & Prose off & 480 & 0 & 0 & 8 & 0 \\
DS & Structured H & 478 & 1 & 28 & 0 & 35 \\
DS & Direct H archive & 288 & 0 & 157 & 0 & 162 \\
GLM & Prose H & 480 & 0 & 85 & 0 & 135 \\
GLM & Prose off & 480 & 0 & 0 & 7 & 0 \\
GLM & Structured H & 479 & 1 & 19 & 0 & 36 \\
GLM & Direct H archive & 288 & 0 & 47 & 1 & 68 \\
\end{longtable}

For each source variant, repeat, and recursive LLM arm, the plan allows
one early and four update writes. A direct archive writer allows one
early and two final branch rewrites. Thus a recursive arm has at most
480 writer operations per backend and direct H at most 288. Across all
conditions the ceiling is 1,728 writer plus 3,456 reader operations per
backend, or 5,184. The count is not obtained by summing writer receipts
copied into several branch rows.

DeepSeek's missing structured reveal writer suppresses a later update
and its reader. GLM has one missing structured override writer and two
missing readers, on prose-off override and prose-H current. The missing
GLM writer suppresses one downstream read. Exported descriptions of
timeout or transmission failure do not replace absent raw error logs.
The single metering-boundary response belongs to an already recorded
DeepSeek prose-H override read, finishes by length, and has no visible
answer. It remains a failed read under the unchanged contract.

The operation totals are consequently DeepSeek 5,181 done + one missing,
and GLM 5,180 done + three missing. Recorded operations need not equal
planned requests, and completed response counts need not equal all
provider-billed attempts. Raw ledgers and invoices are required for the
latter claim. No account prices or missing-call estimates are invented
in the analysis.

\section{Appendix C. Exact diagnostic taxonomy and repair
contract}\label{appendix-c.-exact-diagnostic-taxonomy-and-repair-contract}

For each planned outcome, the classifier first preserves NA as
unresolved. For an observed outcome it attempts to parse the memory as
the public event schema and resolve the gold query. Unparseable memory
is a separate category. Valid but wrong event states are split according
to strict reader success; valid correct states are split into strict
pass, exact-wrapper-only failure, and other reader failure. This creates
a disjoint partition, not independent percentages that can be added
across overlapping writer diagnostics.

The parser and exact answer rules are intentionally conservative. It is
possible for a nonconforming memory to support a correct reader answer.
It is also possible for a malformed reader answer to reveal correct
values. These cases are preserved rather than forced into a single
semantic-loss label. Complete per-row assignments, pair contrasts, and
leave-one-family-out bounds are distributed as CSV files.

Frontier-alpha uses the following strict-memory update. The capitalized
steps refer to the public deterministic event semantics; they are not
calls to a model or an external archive.

\begin{Shaded}
\begin{Highlighting}[]
\NormalTok{ALPHA(previous\_memory, new\_events, targets, budget):}
\NormalTok{  events = NORMALIZE(DECODE(previous\_memory) + new\_events)}
\NormalTok{  live, now = LIVE\_VALUE\_RECORDS\_AND\_MAX\_CLOCK(events)}
\NormalTok{  controls = revocations + last clock attaining now}
\NormalTok{  closure = KEY\_CLOSURE(targets, all live versions)}
\NormalTok{  stable{-}sort controls by is\_clock, descending}
\NormalTok{  stable{-}sort live by (target / closure / other, revision), descending}
\NormalTok{  memory = empty event envelope}
\NormalTok{  for record in controls followed by live:}
\NormalTok{    if canonical UTF{-}8 envelope with record fits budget:}
\NormalTok{      append record}
\NormalTok{  return charged envelope; reject if even the empty envelope cannot fit}
\end{Highlighting}
\end{Shaded}

This tie-break uses the current retained order, not an uncharged
original timestamp or hidden rank. Preserving it through the event array
is necessary for the update contract. If equal-priority records are
permuted, selection may change. If renaming changes escaped
serialization size, capacity may change. Both cases are outside the
proved label-equivariance premise.

Table C1. Offline late-reference budget sweep on the eight original
variants. No model calls are involved; caps were varied retrospectively.
These counts are neither independent test samples nor recommended
optimal capacities.

\begin{longtable}[]{@{}lllllll@{}}
\toprule\noalign{}
Budget (bytes) & 600 & 900 & 1,200 & 1,800 & 2,400 & 4,800 \\
\midrule\noalign{}
\endhead
\bottomrule\noalign{}
\endlastfoot
Original frontier & 8/8 & 8/8 & 8/8 & 8/8 & 8/8 & 8/8 \\
Frontier-alpha & 2/8 & 2/8 & 2/8 & 2/8 & 4/8 & 8/8 \\
\end{longtable}

The repair audit includes 23,328 state-evaluation rows (identity,
renaming, permutation; two selectors; three branches) and 1,728
budget-sweep rows. The 13,440 stage-equivariance comparisons
additionally check intermediate updates, reusing early states where
branches fork. Model-call count and number of new independent tasks are
both zero. No confidence interval is attached to repeated deterministic
transformations.

\section{Appendix D. Reproduction and artifact
identities}\label{appendix-d.-reproduction-and-artifact-identities}

The public-data SHA-256 is

\texttt{ef7984bca91937f31ff6073d3431d0aed7421391e71376a860639cb6eed4a879}.

The gold-data SHA-256 is

\texttt{4bb8f2330378ee4a556691cf044fa50d6057b96bf12a31c30f5e895055c32236}.

The uploaded paid-pilot archive SHA-256 is

\texttt{d64e202c9e53dc5aa5d5e56bb9696e74f81a2ba7d805ce6db03db1e76e7cb610}.

The data seed is 2026091101; run and bootstrap seed is 20260911. The
matching dataset manifest is \texttt{cca4-matched-prefix-v1}, split
\texttt{dev}, with 24 pairs, 48 variants, 32 distractors, and two update
chunks. The distributed reference implementation is labeled as such; it
is not a certification of the entire paid execution tree.

From the package root, \texttt{python\ analysis/run\_\allowbreak{}all.py} regenerates
the numerical analyses without model calls.
\texttt{python\ -m\ unittest\ discover\ -s\ analysis\ -p\ \textquotesingle{}test\_*.py\textquotesingle{}\ -v}
executes unit, metamorphic, accounting, and evidence-boundary tests.
\texttt{python\ analysis/paper\_\allowbreak{}assets.py} rebuilds the tables and
figures from audited CSVs. Building PDF, DOCX, and LaTeX additionally
requires Pandoc and the documented rendering dependencies. The release
manifest verifies the distributed bytes before modification; it is an
integrity check, not a trusted timestamp or third-party attestation.

The artifact includes a claim-to-evidence registry, the old-to-new
revision record, the reference repair, a proposed confirmation contract,
and a source-verification note. The contract is explicitly disabled for
execution and cannot authorize paid experiments. Its purpose is to keep
later empirical claims separate from this exploratory manuscript and to
prevent retrospective development results from being relabeled as
confirmation.

\end{document}